\documentclass{article}

\usepackage{microtype}

\usepackage{subcaption}
\usepackage{booktabs} % for professional tables

\usepackage{hyperref}

\usepackage[preprint]{icml2026}
\usepackage{amsmath}
\usepackage{amssymb}
\usepackage{mathtools}
\usepackage{amsthm}
\usepackage{multirow}
\usepackage{multicol}
\usepackage{xcolor}   % 提供颜色支持
\usepackage{colortbl} % 提供表格行/列颜色支持
\usepackage[capitalize,noabbrev]{cleveref}
\usepackage{graphicx} % 用于 \resizebox
\theoremstyle{plain}

\theoremstyle{definition}

\theoremstyle{remark}

\usepackage[textsize=tiny]{todonotes}

\icmltitlerunning{Efficient Difficulty-Aware Dynamic Routing for Diffusion-Based Real-World Image Super-Resolution}

\begin{document}

\twocolumn[
  \icmltitle{Efficient Difficulty-Aware Dynamic Routing for Diffusion-Based Real-World Image Super-Resolution}



  \icmlsetsymbol{equal}{*}

  \begin{icmlauthorlist}
    % \icmlauthor{Xue Wu}{equal,yyy}
    % \icmlauthor{kang Zhao}{equal,yyy,comp}
    \icmlauthor{Xue Wu}{xd}
    \icmlauthor{Kang Zhao}{ts}
    \icmlauthor{Kafeng Wang}{ts}
    \icmlauthor{Jianfei Chen}{ts} 
    \icmlauthor{Jingwei Xin}{xd}
    \icmlauthor{Nannan Wang}{xd}
    \icmlauthor{Xinbo Gao}{xd}    
  \end{icmlauthorlist}
  
  \icmlaffiliation{xd}{Xidian University, Xi’an 710071, Shaanxi, China.}
  \icmlaffiliation{ts}{Tsinghua University, Beijing, China}
  % \icmlaffiliation{xd}{School of Telecommunications Engineering, Xidian University, Xi’an 710071, Shaanxi, China.}

  \icmlcorrespondingauthor{Nannan Wang}{nnwang@xidian.edu.cn.}

  \icmlkeywords{Image Super-resolution}

  \vskip 0.3in
]



\printAffiliationsAndNotice{}  
% no special notice (required even if empty)

\begin{abstract}
  
   Diffusion-based methods have achieved impressive performance in real-world image super-resolution (Real-ISR) by leveraging large pre-trained stable diffusion (SD) models as powerful generative priors. However, these methods still face two key limitations. First, existing SD-based one-step and multi-step Real-ISR approaches adopt a unified processing paradigm for all input samples, ignoring the varying restoration difficulty across images. Second, the aggressive resolution reduction of the VAE in SD models (e.g., 8× downsampling) leads to irreversible loss of fine-scale details, which cannot be recovered by the subsequent diffusion process. 
  To address these limitations, we propose a Difficulty-aware Dynamic Routing (DDR) strategy that overcomes the rigid, one-size-fits-all processing paradigm. Specifically, we first design a difficulty estimator to predict the restoration cost of each input image, enabling automatic assignment to a network of appropriate capacity. Then, we construct a set of Real-ISR networks with varying model capacities by modulating the spatial downsampling ratio of the VAE in the SD backbone, thereby preserving more high-frequency information for challenging cases while maintaining efficiency for simpler inputs.
  Extensive experiments have demonstrated the superior efficiency and effectiveness of the proposed model compared to recent state-of-the-art methods.
  % Therefore, diffusion models should reason for each instance to adaptively determine the optimal noise schedule, achieving high generation quality with sampling efficiency. With such an adaptive scheduler, TPDM not only generates high-quality images that are aligned closely with human preferences but also adjusts diffusion time and the number of denoising steps on the fly, enhancing both performance and efficiency. With Stable Diffusion 3 Medium architecture, TPDM achieves an aesthetic score of 5.44 and a human preference score (HPS) of 29.59, while using around 50 fewer denoising steps to achieve better performance.

\end{abstract}

\section{Introduction}

% kafeng add from deepseek
\begin{figure}[ht]
  \centering
  \includegraphics[width=\columnwidth]{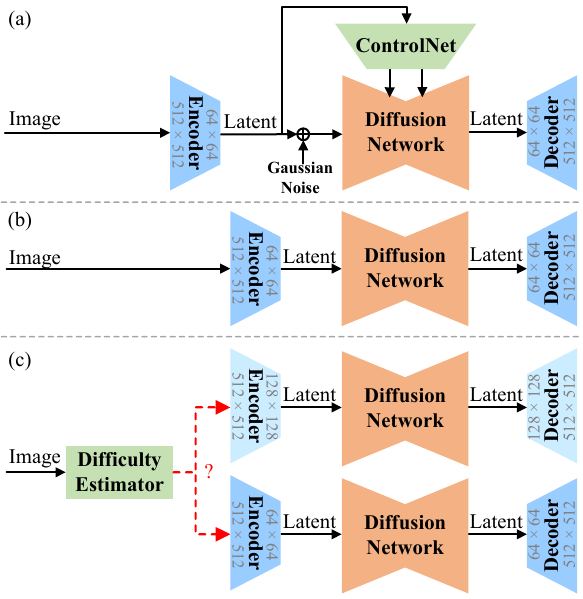}
  \caption{
    Architecture comparison of diffusion-based SR approaches.
    (a) Controller-based methods employ conditional mechanisms to guide the
    diffusion process; (b) Standard diffusion processes latent representations
    directly; (c) Our proposed DDR employs difficulty-aware dynamic routing to
    adaptively select expert networks with varying VAE compression ratios.
  }
  \label{fig1}
  \vspace{-0.2in}
\end{figure}
Single image super-resolution (SR) \cite{wang2020deep} is a classical yet still active low-level vision problem, aiming to reconstruct a high-resolution (HR) image from its low-resolution (LR) counterpart, which suffers from unknown degradations. SR is also a challenging ill-posed problem, as multiple plausible HR images may correspond to the same LR input. This problem has attracted long-standing and widespread attention in the computer vision community \cite{yang2010image, dong2015image}.

Over the past decade, numerous seminal studies have been conducted to tackle this challenge, utilizing convolutional Neural Networks (CNNs) \cite{dong2016accelerating, kim2016accurate, lim2017enhanced}, vision transformers (ViTs) \cite{liang2021swinir, chen2023hat}, and their combinations \cite{chen2023dual}. Despite the remarkable progress achieved, these methods mostly fail in real-world scenarios. The failure stems from the fact that these traditional SR approaches assume a known degradation process, thereby failing to account for the unknown and complex degradations present in real-world low-quality images \cite{wang2021real, zhang2021designing}. Consequently, real-world image super-resolution (Real-ISR) is designed to reconstruct perceptually realistic HR images in real-world scenarios, thereby offering greater practical utility \cite{cai2019realworld}. In recent years, Real-ISR has attracted increasing attention from researchers. Many researchers have focused on the use of generative models, particularly generative adversarial networks (GANs), for Real-ISR, owing to their ability to recover sharp and photorealistic textures from low-quality inputs via adversarial training and efficient one-step inference \cite{ledig2017photo, wang2021realesrgan}. However, GAN-based methods suffer from inherent limitations, including training instability, mode collapse, and a limited generative prior, which often result in hallucinated details and perceptually unpleasant artifacts under complex real-world degradations \cite{goodfellow2020gan}.

Recently, diffusion models (DMs) have emerged as a powerful class of generative models and have shown impressive performance in image generation as well as many downstream tasks such as Real-ISR. Their robust priors empower them to produce more realistic images with richer details than GAN-based methods \cite{ho2020denoising, rombach2022high, saharia2022photorealistic}. Based on pre-trained stable diffusion (SD) models \cite{rombach2022high}, many Real-ISR methods \cite{wang2023stablesr, yang2023pixel, yu2024scaling, wu2024seesr, lin2023diffbir} have been proposed to improve the realistic details of SR outputs. Representative works such as StableSR \cite{wang2023stablesr}, PASD \cite{yang2023pixel}, SUPIR \cite{yu2024scaling}, and SeeSR \cite{wu2024seesr} take the LR input as the control signal and utilize ControlNet \cite{zhang2023adding} to guide the SR process.
However, by typically initializing the diffusion process with random noise, these methods not only introduce unwanted randomness into the output images but also require multiple denoising steps to reconstruct the final high-quality result, rendering the Real-ISR process computationally expensive \cite{song2020denoising}. Very recently, to accelerate the generation process of Real-ISR, one-step SD-based methods \cite{wang2023sinsr, xie2024addsr, li2024distillation, yin2024one} have been developed. These methods directly take LR images as input and use distillation techniques to fine-tune the denoising process, presenting a promising approach for efficient and realistic super-resolution under real-world degradations.

Despite these advancements, our investigation has revealed two critical limitations associated with existing DM-based Real-ISR methods. Firstly, existing DM-based Real-ISR methods predominantly focus on architectural design and training objective optimization. As shown in \Cref{fig1}(a) and (b), once trained, these models are typically applied in a uniform manner to all input samples, regardless of their intrinsic degradation characteristics or restoration difficulty. However, we observe that real-world low-quality images exhibit significant variability in degradation complexity, and little attention has been paid to dynamically allocating different SR networks or inference strategies according to the difficulty of the input samples. This limitation motivates the exploration of adaptive Real-ISR frameworks that can more efficiently and effectively handle diverse real-world degradations \cite{kong2021classsr, liu2022adaptive}. Secondly, the aforementioned multi-step and one-step SD-based approaches encode LR inputs into a latent space using a pretrained variational autoencoder (VAE) \cite{rombach2022high} and either directly feed these latents into the denoising UNet or use them to condition the denoising process through a controller mechanism. However, the VAE’s aggressive downsampling (e.g., 8× in SD 2.1-base \cite{rombach2022high}) inevitably discards high-frequency details \cite{zhu2023designing}. Moreover, since VAEs are trained on high-quality images, encoding low-quality LR inputs often degrades structural fidelity \cite{yi2025fine}. Consequently, these methods struggle to reconstruct complex textures and fine patterns.

To address these limitations, we propose an efficient SD-based Difficulty-aware Dynamic Routing (DDR) approach, coined as DDR-SR, which can adaptively assign an appropriate inference model to each sample based on its restoration difficulty, thereby achieving fine-structure-preserving Real-ISR while balancing reconstruction performance and efficiency. \Cref{fig1}(c) illustrates the overall framework. In image restoration tasks, the recovery difficulty of an input image is primarily determined by the degree of high-frequency structural information loss. Accordingly, we adopt high-frequency energy attenuation as a quantitative metric to assess its restoration difficulty \cite{ma2020structure}.
To enable targeted processing of inputs with varying complexity, we design a set of specialized expert networks, each tailored to a specific range of image difficulty. Specifically, challenging samples are routed to a model equipped with a VAE featuring a low spatial compression ratio, which preserves richer fine-grained details in the latent space \cite{yi2025fine}. In contrast, easier samples are assigned to a model using a VAE with a higher compression ratio, prioritizing inference efficiency without compromising perceptual quality. This dynamic allocation strategy not only avoids over-processing simple inputs but also ensures that complex images receive sufficient modeling capacity, leading to a more optimal trade-off between visual fidelity and computational cost. Moreover, the flexible design of DDR allows users to explicitly control the trade-off between visual fidelity and inference efficiency by adjusting the routing ratio between high- and low-capacity expert networks, thereby enabling application-specific customization.
Extensive experiments demonstrate that our DDR method outperforms state-of-the-art SD-based Real-ISR methods in achieving a better trade-off between reconstruction performance and inference efficiency.
In summary, our contributions are as follows:
\begin{itemize}
  \item We propose DDR-SR, the first difficulty-aware dynamic routing framework for SD-based Real-ISR, which moves beyond the conventional one-size-fits-all paradigm by adaptively selecting inference models based on restoration difficulty to enable sample-adaptive processing.
  \item We design a set of VAE-modulated expert networks with varying spatial compression ratios, which jointly preserve fine structural details for challenging images and ensure efficient restoration for simpler inputs.
  \item Extensive experiments demonstrate that DDR achieves a superior balance between reconstruction performance and inference efficiency, outperforming state-of-the-art SD-based Real-ISR approaches while offering flexible customization for diverse application scenarios.
\end{itemize}

\section{Related Work}
% kafeng add from deepseek
\subsection{Multi-step Diffusion-based Real-ISR}
Diffusion models have revolutionized generative image modeling with their robust priors and high-quality outputs \cite{ho2020denoising, rombach2022high}. A significant line of research leverages pre-trained Text-to-Image (T2I) diffusion models, such as Stable Diffusion (SD) \cite{rombach2022high}, as powerful generative priors for the Real-ISR task. These methods typically use the Low-Resolution (LR) image as a conditioning signal to guide a multi-step iterative denoising process, achieving impressive perceptual quality.
StableSR \cite{wang2023stablesr} fine-tunes a time-aware encoder and employs controllable feature wrapping modules to balance fidelity and realism. DiffBIR \cite{lin2023diffbir} adopts a two-stage strategy, first using a reconstruction network for preliminary restoration and then employing SD for detail enhancement. To better harness the generative power of diffusion models, several works introduce more sophisticated conditioning mechanisms. SeeSR \cite{wu2024seesr} extracts semantic information from the image to provide high-level guidance. PASD \cite{yang2023pixel} introduces a pixel-aware cross-attention module for local structure perception and a degradation removal module to extract robust features. SUPIR \cite{yu2024scaling} scales up model and data size, utilizing negative prompts and restoration-guided sampling to achieve state-of-the-art generative and fidelity capabilities. Recent works continue to explore new frontiers: DiffSteISR \cite{diffsteisr} extends diffusion prior to stereo image super-resolution, addressing view consistency, while EPDiff \cite{epdiff} and ConsisSR \cite{consissr} focus on integrating stronger semantic priors and improving pixel-level consistency, respectively. However, a fundamental limitation of all these methods is their reliance on iterative denoising (often 20-50 steps), resulting in high computational cost and slow inference, which hinders practical deployment \cite{li2024distillation}.

\subsection{One-step Diffusion-based Real-ISR}
To overcome the speed bottleneck of multi-step diffusion, there is a growing interest in distilling these models into efficient one-step generators \cite{yin2023one, liu2023instaflow, salimans2022progressive}. This line of work aims to achieve real-time or near-real-time Real-ISR while preserving generative quality.
SinSR \cite{wang2023sinsr} applies consistency distillation to compress the multi-step ResShift \cite{yue2024resshift} model into a single step, though its generalization is limited by training data scale. AddSR \cite{xie2024addsr} introduces Adversarial Diffusion Distillation (ADD) \cite{sauer2025adversarial} to Real-ISR, producing a fast four-step model that can sometimes generate excessive details. OSEDiff \cite{li2024distillation} proposes a more direct approach by using the LR image as the starting point of the diffusion process (instead of random noise) and employs Variational Score Distillation (VSD) loss \cite{wang2024prolificdreamer} for regularization. To address the common issue of spurious details in one-step models, StructSR \cite{li2025structsr} proposes a plug-and-play, inference-time mechanism that suppresses artifacts and enhances structural fidelity. Beyond single images, the one-step paradigm is also being explored for video; DLoRAL \cite{dloral} employs a dual-LoRA design to decouple temporal consistency and spatial detail learning for efficient video super-resolution. While these methods significantly improve speed, they still operate under a static, ``one-size-fits-all'' inference paradigm, applying the same network capacity to all inputs regardless of their restoration difficulty \cite{wu2024seesr}.

\section{Methodology}
% ---------------------------------Fig2--------------------------------
\subsection{Motivation}
\begin{figure}[ht]
  \begin{center}
    \centerline{\includegraphics[width=\columnwidth]{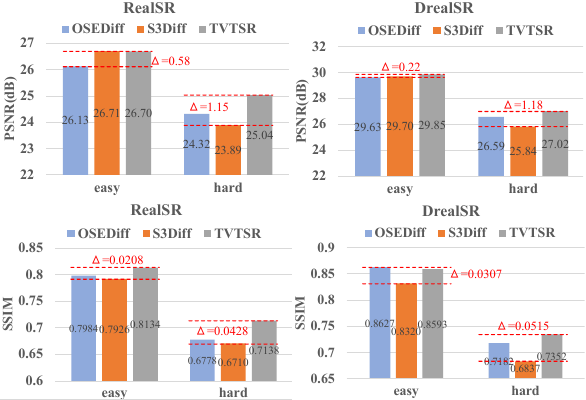}}
    \caption{
     The PSNR and SSIM performance of three representative SOTA methods (OSEDiff \cite{wu2024one}, S3Diff\cite{zhang2024degradation}, and TVTSR\cite{yi2025fine}) on both easy and hard subsets. Notably, the performance gaps among these methods are consistently smaller on the easy subset than on the hard subset across both metrics.
    }
    \label{fig2}
  \end{center}
  \vskip -0.2in
\end{figure}
% ---------------------------------Fig2--------------------------------
% ---------------------------------Fig3--------------------------------

\begin{figure*}[!t]
  \begin{center}
    \centerline{\includegraphics[width=\linewidth]{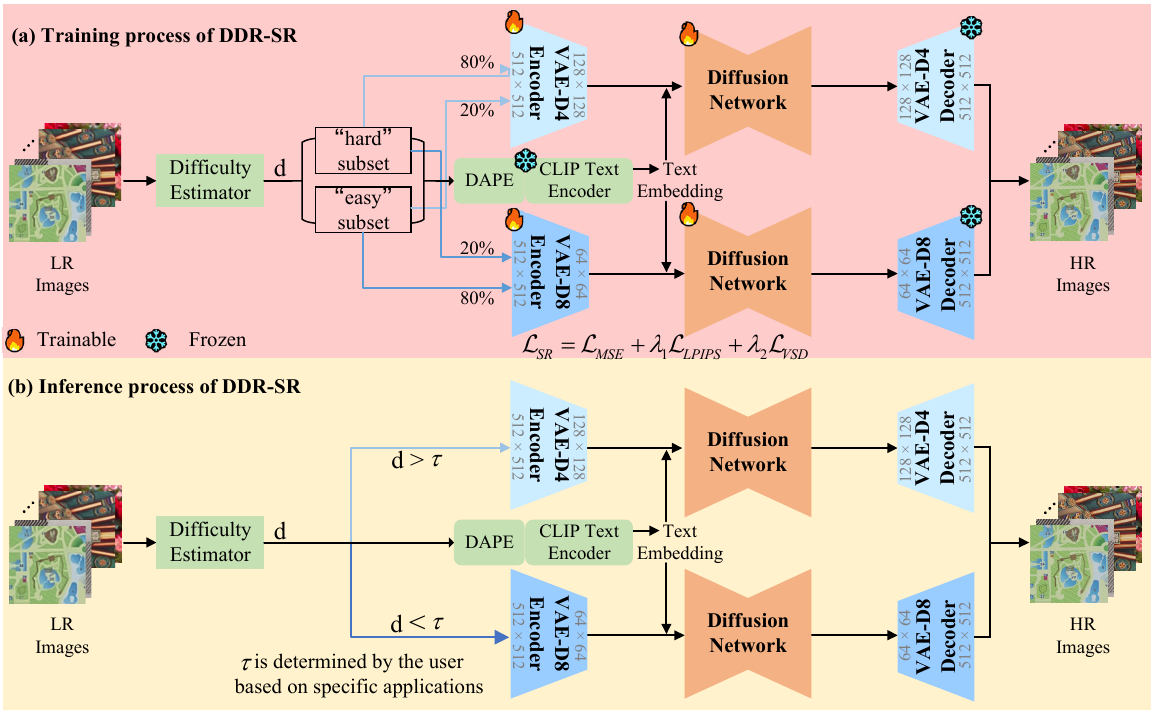}}
    \caption{
     Overview of the proposed Difficulty-aware Dynamic Routing (DDR) framework. During training, the dataset is split into “easy” and “hard” subsets using a high-frequency energy–based difficulty estimator. We train two expert models via LoRA: VAE-D4 + diffusion network on 80\% hard + 20\% easy samples for detail preservation, and VAE-D8 + diffusion network on 80\% easy + 20\% hard samples for efficiency. The two networks are trained with a combination of a reconstruction loss, an LPIPS loss, and a VSD loss. At inference, users can explicitly control the trade-off between visual fidelity and speed by adjusting the routing ratio between the high-capacity (D4) and low-capacity (D8) experts.
    }
    \label{fig3}
  \end{center}
  \vskip -0.2in
\end{figure*}
% ---------------------------------Fig3--------------------------------
To better understand the impact of input difficulty on existing methods, we conduct a systematic evaluation of state-of-the-art one-step SD-based Real-ISR approaches across samples of varying restoration complexity. Using our high-frequency energy attenuation metric, we estimate the difficulty of all images in the RealSR \cite{realsr} and DRealSR \cite{drealsr} benchmarks and split each dataset into “easy” and “hard” subsets based on the median difficulty score. 
As shown in \Cref{fig2}, we compare the PSNR and SSIM performance of three representative SOTA methods (OSEDiff \cite{wu2024one}, S3Diff\cite{zhang2024degradation}, and TVTSR\cite{yi2025fine}) on both subsets.
Strikingly, we find that while performance gaps among competing methods are relatively small on the easy subset, indicating that most models can adequately handle mild degradations, these gaps become significantly pronounced on the hard subset, where fine structural details are severely degraded. This observation reveals a critical shortcoming: current one-size-fits-all frameworks lack the capacity to adapt their inference strategy to the intrinsic complexity of each input, leading to suboptimal reconstruction of challenging samples. Motivated by this insight, we argue that an effective Real-ISR system should not treat all inputs equally, but rather allocate computational resources and modeling capacity in a difficulty-aware manner. This motivates our design of a dynamic routing mechanism that tailors the inference path to the restoration difficulty of each sample.

\subsection{Difficulty Estimator}
\label{sec:difficulty_estimator}
In image restoration tasks, the recovery difficulty is largely determined by the amount of high-frequency structural information lost during degradation, as high-frequency components correspond to edges and fine details that are critical yet difficult to reconstruct. Motivated by this observation, we quantify restoration difficulty using high-frequency energy attenuation by comparing the high-frequency spectral energy of the low-quality image with that of its ground-truth counterpart. This metric directly reflects the severity of detail loss caused by degradation and aligns well with the physical characteristics of real-world imaging processes, which predominantly act as low-pass filters. Moreover, it is model-agnostic, computationally efficient, and thus suitable for large-scale dataset analysis.

Formally, we define the restoration difficulty $d$ of a degraded image $I_{deg}$ relative to its ground-truth counterpart $I_{gt}$ as the normalized loss of high-frequency spectral energy induced by the degradation process. Specifically, let $\mathcal{F}(\cdot)$ denote the 2D Fourier transform, and let $\mathcal{M}_{high}$ be a binary mask that selects high-frequency regions in the frequency domain (e.g., coefficients beyond a radial threshold). The high-frequency energy of an image $I$ is then computed as:
\begin{equation}
E_{\text{high}}(I) = \sum_{(u,v) \in \mathcal{M}_{\text{high}}} \big| \mathcal{F}(I)(u,v) \big|^2.
\label{eq:1}
\end{equation}
The restoration difficulty of a degraded image is defined as:
\begin{equation}
d = 1 - \frac{E_{\text{high}}(I_{\text{deg}})}{E_{\text{high}}(I_{\text{gt}})}.
\label{eq:2}
\end{equation}
This formulation yields a value in the range  $ [0, 1] $, where  $ d = 0 $  indicates no high-frequency loss (trivial to restore), and  $ d = 1 $  implies complete suppression of high-frequency content (maximally challenging). By design,  $ d $  captures the intrinsic complexity of restoring fine structures from the degraded input, providing a reliable and interpretable signal for dynamic routing in our framework.

\subsection{Training of DDR-SR}
 Existing SD-based one-step and multi-step Real-ISR methods adopt a one-size-fits-all paradigm, which lacks the ability to adapt their inference strategy to the intrinsic complexity of each input. This limitation leads to suboptimal reconstruction of challenging samples with severe degradation, while simultaneously wasting computational resources on easy inputs that require less modeling capacity. To overcome this fundamental drawback, our goal is to move beyond the uniform processing paradigm and build an effective Real-ISR system that does not treat all inputs equally. Instead, DDR-SR allocates computational resources and modeling capacity in a difficulty-aware manner, ensuring that each sample receives an appropriate level of processing tailored to its restoration difficulty. 
 
To realize this vision, we first estimate the restoration difficulty of each training sample using the high-frequency energy attenuation metric introduced in Section \ref{sec:difficulty_estimator}. Based on the estimated difficulty scores, the entire training set is partitioned into ``easy'' and ``hard'' subsets. We then construct two specialized expert networks, denoted as Expert-D4 and Expert-D8, which share the same diffusion UNet backbone but employ VAEs with different spatial compression ratios (4$\times$ and 8$\times$, respectively). Specifically, we adopt the VAE-D4 from \cite{yi2025fine} as the VAE in Expert-D4 due to its superior detail preservation capability, and use the VAE from SD 2.1-base as the VAE in Expert-D8.
Crucially, rather than training each expert exclusively on its designated subset, we adopt a mixed-training strategy to enhance robustness and generalization. Specifically, Expert-D4, designed for high-fidelity reconstruction, is trained on a mixture of 80\% hard samples and 20\% easy samples, encouraging it to focus on complex cases while retaining baseline capability on simpler inputs. Conversely, Expert-D8, optimized for inference efficiency, is trained on 80\% easy samples and 20\% hard samples, allowing it to handle common degradations efficiently without completely failing on challenging instances.
Both experts are fine-tuned from a pre-trained SD using Low-Rank Adaptation (LoRA) \cite{hu2021lora}, applied to the VAE encoder and the weights of all convolution and MLP layers of the diffusion UNet. This parameter-efficient adaptation enables each expert to develop distinct sensitivity to degradation patterns corresponding to their target difficulty range, while preserving the rich prior knowledge of the base model. The resulting experts thus exhibit complementary strengths: Expert-D4 excels at recovering fine structures in severely degraded images, whereas Expert-D8 achieves fast and perceptually satisfactory results on mildly degraded inputs.

\begin{algorithm}[t]
\caption{DDR-SR Training Procedure}
\label{alg:ddr_training}
\begin{algorithmic}[1]
\REQUIRE Training dataset $\mathcal{D} = \{ (I_{\text{deg}}^{(i)}, I_{\text{gt}}^{(i)}) \}_{i=1}^N$;  
         Difficulty estimator $\mathcal{E}_{\text{diff}}(\cdot)$;  
         VAE compression ratios $r_4 = 4$, $r_8 = 8$;  
         Mixing ratios $\alpha = 0.8$, $\beta = 0.2$.
\ENSURE Trained expert models: Expert-D4, Expert-D8.

\STATE // \textbf{Step 1: Estimate difficulty and split dataset}
\FOR{each $(I_{\text{deg}}, I_{\text{gt}}) \in \mathcal{D}$}
    \STATE Compute difficulty score: $d \gets \mathcal{E}_{\text{diff}}(I_{\text{deg}}, I_{\text{gt}})$
\ENDFOR
\STATE Split $\mathcal{D}$ into easy subset $\mathcal{D}_{\text{easy}}$ and hard subset $\mathcal{D}_{\text{hard}}$ using median $d$.

\STATE // \textbf{Step 2: Construct mixed training sets}
\STATE $\mathcal{D}_{\text{D4}} \gets \alpha \cdot \mathcal{D}_{\text{hard}} + \beta \cdot \mathcal{D}_{\text{easy}}$  \hfill // for high-fidelity expert
\STATE $\mathcal{D}_{\text{D8}} \gets \alpha \cdot \mathcal{D}_{\text{easy}} + \beta \cdot \mathcal{D}_{\text{hard}}$  \hfill // for efficient expert

\STATE // \textbf{Step 3: Initialize experts from pre-trained SD}
\STATE Initialize Expert-D4 with VAE ($r_4$) + UNet
\STATE Initialize Expert-D8 with VAE ($r_8$) + UNet
\STATE Apply LoRA to the VAE encoder and UNet in both experts.

\STATE // \textbf{Step 4: Fine-tune experts}
\FOR{epoch = 1 to $T$}
    \FOR{batch $\mathcal{B}_{\text{D4}} \sim \mathcal{D}_{\text{D4}}$}
        \STATE Update Expert-D4 via diffusion loss $\mathcal{L}_{\text{SR}}$
    \ENDFOR
    \FOR{batch $\mathcal{B}_{\text{D8}} \sim \mathcal{D}_{\text{D8}}$}
        \STATE Update Expert-D8 via diffusion loss $\mathcal{L}_{\text{SR}}$
    \ENDFOR
\ENDFOR

\STATE \textbf{Return} Expert-D4, Expert-D8
\end{algorithmic}
\end{algorithm}
\Cref{fig3}(a) illustrates the training process of our approach. Taking the training of Expert-D8 as an example, after dividing the dataset into easy and hard subsets based on their restoration difficulty, a degraded image $I_{deg}$ is fed into the Expert-D8. We encode it into a latent feature ${z_{\deg }} = {E_8}({I_{\deg }})$, and extract the text embedding $c_{t}$ from it by a prompt extractor, which includes the DAPE \cite{wu2024seesr} and CLIP text encoder \cite{rombach2022high}. The restored latent feature $z_{SR}$ is obtained by passing $z_{\deg }$ and $c_{t}$ through the denoising UNet ${\epsilon_8}$:
\begin{equation}
{z_{SR}} = \frac{{z - \sqrt {1 - {{\bar \alpha }_t}} {\epsilon_8}(z,{c_t},t)}}{{\sqrt {{{\bar \alpha }_t}} }},
\label{eq:3}
\end{equation}
where the time step $t$ is set to 1 and ${\bar \alpha }_t$ is the corresponding diffusion coefficient \cite{ho2020denoising}. Finally, feeding $z_{SR}$ into the decoder $D_8$, we obtain the SR image $I_{SR}$:
\begin{equation}
{I_{SR}} = {D_8}(z_{SR}).
\label{eq:4}
\end{equation}
To achieve Real-ISR with one-step diffusion, following OSEDiff \cite{wu2024one}, we use $\mathcal{L}_{MSE}$ loss, LPIPS loss and VSD loss to train our Expert-D4 and Expert-D8:
\begin{equation}
{\mathcal{L}_{SR}} = \mathcal{L}_{MSE} + {\lambda _1}{\mathcal{L}_{LPIPS}} + {\lambda _2}{\mathcal{L}_{VSD}}.
\label{eq:5}
\end{equation}
where ${\lambda _1}=2$, ${\lambda _2}=1$ are weighting hyper-parameters.
In summary, the overall training procedure of DDR-SR is presented in \cref{alg:ddr_training}.
% \begin{algorithm}[tb]
%   \caption{DDR-SR Training Procedure}
%   \label{alg:example}
%   \begin{algorithmic}
%     \STATE {\bfseries Input:} data $x_i$, size $m$
%     \REPEAT
%     \STATE Initialize $noChange = true$.
%     \FOR{$i=1$ {\bfseries to} $m-1$}
%     \IF{$x_i > x_{i+1}$}
%     \STATE Swap $x_i$ and $x_{i+1}$
%     \STATE $noChange = false$
%     \ENDIF
%     \ENDFOR
%     \UNTIL{$noChange$ is $true$}
%   \end{algorithmic}
% \end{algorithm}

\subsection{The Inference Process of DDR-SR}
During inference, DDR-SR operates in a fully automatic and sample-adaptive manner. Given a low-quality input image $I_{deg}$, the system first estimates its restoration difficulty using the difficulty estimator (as defined in Section \ref{sec:difficulty_estimator}). This yields a scalar difficulty score $d \in [0,1]$, where higher values indicate more severe high-frequency loss and greater reconstruction challenge.

Based on $d$, a dynamic router assigns the input to one of two pre-trained expert networks:
\begin{itemize}
    \item If $d \geq \tau$, the sample is routed to \textbf{Expert-D4}, which employs a VAE with a low spatial compression ratio (4$\times$) to preserve fine-grained details in the latent space, prioritizing visual fidelity.
    \item Otherwise, it is processed by \textbf{Expert-D8}, which uses a higher compression ratio (8$\times$) VAE for faster inference while maintaining perceptual quality on mildly degraded inputs.
\end{itemize}

The threshold $\tau$ controls the routing balance between the two experts. By default, we set $\tau$ to the median difficulty of the training set, yielding an approximately 50/50 split. However, a key advantage of DDR-SR is its \textit{user-controllable trade-off}: practitioners can explicitly adjust $\tau$ (or directly specify the desired routing ratio) to favor either reconstruction quality or computational efficiency according to application needs. The inference process of DDR-SR is presented in \cref{alg:ddr_inference}.
% For instance, in latency-sensitive scenarios (e.g., mobile deployment), one may lower $\tau$ to route more samples through Expert-D8; conversely, for high-fidelity applications (e.g., medical or satellite imaging), $\tau$ can be raised to prioritize Expert-D4.

% Notably, the entire pipeline—including difficulty estimation, routing, and expert inference—requires no ground-truth reference and runs efficiently on a single GPU. This makes DDR-SR practical for real-world deployment while delivering adaptive super-resolution performance tailored to each input’s intrinsic complexity.
\begin{algorithm}[h]
\caption{Inference Process of DDR-SR}
\label{alg:ddr_inference}
\begin{algorithmic}[1]
\REQUIRE Degraded input image $I_{\text{deg}}$;  
         Difficulty estimator $\mathcal{E}_{\text{diff}}(\cdot)$;  
         Trained experts: Expert-D4, Expert-D8;  
         Routing threshold $\tau \in [0,1]$ (default: median of training difficulties).
\ENSURE High-resolution output $I_{\text{hr}}$.

\STATE // \textbf{Step 1: Estimate restoration difficulty}
\STATE Compute difficulty score: $d \gets \mathcal{E}_{\text{diff}}(I_{\text{deg}}, I_{\text{gt}})$

\STATE // \textbf{Step 2: Dynamic routing based on difficulty}
\IF{$d \geq \tau$}
    \STATE // Route to high-fidelity expert
    \STATE $I_{\text{SR}} \gets \text{Expert-D4}(I_{\text{deg}})$
\ELSE
    \STATE // Route to efficient expert
    \STATE $I_{\text{SR}} \gets \text{Expert-D8}(I_{\text{deg}})$
\ENDIF

\STATE \textbf{Return} $I_{\text{SR}}$
\end{algorithmic}
\end{algorithm}

\begin{table*}[t]
\caption{Quantitative comparison with state-of-the-art multi-step and one-step methods across both a synthetic and two real-world benchmarks.
“-N” behind the method name represents the number of inference steps. Horizontal line is used to separate multi-step and one-step methods, which improves the readability of the table. The best and second best results of each metric are highlighted in \textcolor{red}{red} and \textcolor{blue}{blue}, respectively.}
    \label{tab:1}
    \centering
    % \footnotesize
    % \scriptsize{}
    \setlength{\tabcolsep}{3pt}
    \renewcommand{\arraystretch}{1.2}
    \begin{tabular}{c|c|ccccccccc}
        \hline
        \multirow{2}{*}{Datasets} & \multirow{2}{*}{Methods} & \multicolumn{9}{c}{Metrics} \\
        \cline{3-11}
         &  & PSNR$\uparrow$ & SSIM$\uparrow$ & LPIPS$\downarrow$ & DISTS$\downarrow$ & FID$\downarrow$ &CLIPIQA$\uparrow$ & MUSIQ$\uparrow$ & MANIQA$\uparrow$ & NIQE$\downarrow$ \\
        \hline
        \multirow{8}{*}{DIV2K-val}
        % & \multicolumn{2}{l}{\textcolor{gray}{Multi-step Diffusion Methods}}\\
        % \cline{2-11}
        & StableSR-200 &23.31 &0.5728 &0.3129 &0.2138 &\textcolor{red}{24.67} &0.6682  &65.63  &0.6188  &4.76  \\
        & DiffBIR-50   &23.67 &0.5653 &0.3541 &0.2129 &30.93 &0.6652  &65.66  &0.6204  &\textcolor{blue}{4.71}  \\
        & SeeSR-50       &23.71 &0.6045 &0.3207 &\textcolor{blue}{0.1967} &25.83 &\textcolor{blue}{0.6857}  &68.49 &\textcolor{blue}{0.6239} &4.82  \\
        & PASD-20         &23.14 &0.5489 &0.3607 &0.2219 &29.32 &0.6711 &\textcolor{red}{68.83} &\textcolor{red}{0.6484} &\textcolor{red}{4.40}  \\
        & ResShift-15     &\textcolor{red}{24.69} &\textcolor{blue}{0.6175} &0.3374 &0.2215 &36.01 &0.6089 &60.92 &0.5450  &6.82  \\
        \cline{2-11}
        % & \multicolumn{2}{l}{\textcolor{gray}{One-step Diffusion Methods}} \\
        % \cline{2-11}
        & SinSR-1       &\textcolor{blue}{24.43} &0.6012 &0.3262 &0.2066 &35.45 &0.6499 &62.80 &0.5395 &6.02  \\
        & OSEDiff-1      &23.72 &0.6108 &\textcolor{blue}{0.2941} &0.1976 &26.32 &0.6683 &67.97 &0.6148 &\textcolor{blue}{4.71}  \\
        \rowcolor{red!10}
        & DDR-SR-1       &24.09 &\textcolor{red}{0.6257} &\textcolor{red}{0.2804} &\textcolor{blue}{0.1911} &\textcolor{blue}{25.13} &\textcolor{red}{0.6908} &\textcolor{blue}{68.66}  &0.6230  &5.02  \\
        \hline
        \multirow{8}{*}{RealSR}
        % & \multicolumn{2}{l}{\textcolor{gray}{Multi-step Diffusion Methods}}\\
        % \cline{2-11}
        & StableSR-200 &24.70 &0.7085 &0.3018 &0.2167 &127.20 & 0.6178 & 65.78 & 0.6221 & 5.9122 \\
        & DiffBIR-50   &24.75 &0.6567 &0.3636 &0.2290 &124.56 & 0.6463 & 64.98 & 0.6246 & 5.5346 \\
        & SeeSR-50       &25.18 &0.7216 &0.3009 &0.2213 &125.66 & 0.6612 & \textcolor{red}{69.77} & \textcolor{blue}{0.6442} & \textcolor{red}{5.4081} \\
        & PASD-20         &25.21 &0.6798 &0.3380 &0.2259 &\textcolor{blue}{123.08} & 0.6620 & 68.75 & \textcolor{red}{0.6487} & \textcolor{blue}{5.4137} \\
        & ResShift-15 &\textcolor{red}{26.31} &\textcolor{blue}{0.7421} &0.3460 &0.2498 &142.81 & 0.5444 & 58.43 & 0.5285 & 7.2635 \\
        \cline{2-11}
        % & \multicolumn{2}{l}{\textcolor{gray}{One-step Diffusion Methods}} \\
        % \cline{2-11}
        & SinSR-1       &\textcolor{blue}{26.28} &0.7347 &0.3188 &0.2346 &137.05 & 0.6122 & 60.80 & 0.5385 & 6.2872 \\
        & OSEDiff-1      &25.15 &0.7341 &\textcolor{blue}{0.2921} &\textcolor{blue}{0.2128} &123.50 & \textcolor{blue}{0.6693} & 69.09 & 0.6326 & 5.6476 \\
        \rowcolor{red!10}
        & DDR-SR-1       &25.92 &\textcolor{red}{0.7555} &\textcolor{red}{0.2634} &\textcolor{red}{0.2044} &\textcolor{red}{114.86} & \textcolor{red}{0.6725} & \textcolor{blue}{69.32} & 0.6429 & 5.8048 \\
        \hline
        \multirow{8}{*}{DRealSR}
        % & \multicolumn{2}{l}{\textcolor{gray}{Multi-step Diffusion Methods}}\\
        % \cline{2-11}
        & StableSR-200 &28.03 &0.7536 &0.3284 &0.2287 &147.03 & 0.6356 & 58.51 & 0.5601 & 6.5239 \\
        & DiffBIR-50   &26.71 &0.6571 &0.4557 &0.2706 &167.38 & 0.6395 & 61.07 & 0.5930 & \textcolor{blue}{6.3124} \\
        & SeeSR-50       &28.17 &0.7691 &0.3189 &0.2306 &149.86 & 0.6804 & \textcolor{blue}{64.93} & \textcolor{blue}{0.6042} & 6.3967 \\
        & PASD-20        &27.36 &0.7073 &0.3760 &0.2535 &157.36 & 0.6808 & 64.87 & \textcolor{red}{0.6169} & \textcolor{red}{5.5474} \\
        & ResShift-15 &\textcolor{red}{28.46} &0.7673 &0.4006 &0.2700 &175.92 & 0.5342 & 50.60 & 0.4586 & 8.1249 \\
        \cline{2-11}
        % & \multicolumn{2}{l}{\textcolor{gray}{One-step Diffusion Methods}}\\
        % \cline{2-11}
        & SinSR-1       &28.36 &0.7515 &0.3665 &0.2488 &177.05 & 0.6383 & 55.33 & 0.4884 & 6.9907 \\
        & OSEDiff-1      &27.92 &\textcolor{blue}{0.7835} &\textcolor{blue}{0.2968} &\textcolor{red}{0.2165} &\textcolor{red}{135.29}  & \textcolor{blue}{0.6963} & 64.65 & 0.5899 & 6.4902 \\
        \rowcolor{red!10}
        & DDR-SR-1         &\textcolor{blue}{28.43} &\textcolor{red}{0.7945} &\textcolor{red}{0.2902} &\textcolor{blue}{0.2191} &\textcolor{blue}{137.82} & \textcolor{red}{0.7084} & \textcolor{red}{65.98} & 0.6012 & 6.7650 \\
        \hline
    \end{tabular}
\end{table*}
\begin{figure*}[!t]
  \begin{center}
    \centerline{\includegraphics[width=\linewidth]{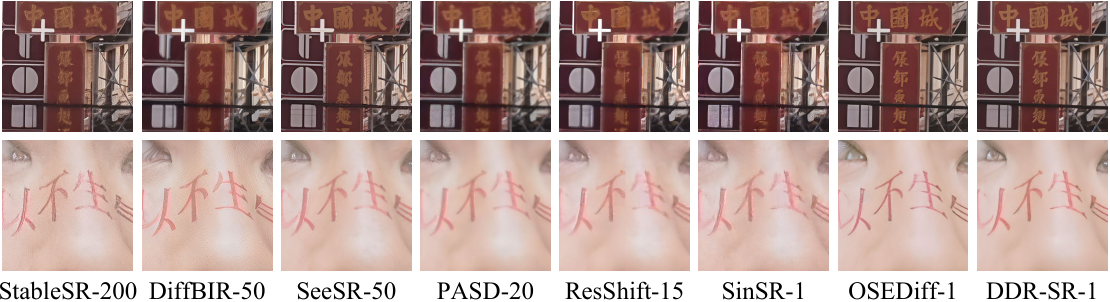}}
    \caption{
    Visual comparisons of different Real-ISR methods. Please zoom in for a better view.
    }
    \label{fig4}
  \end{center}
  \vskip -0.2in
\end{figure*}

\section{Experiments}

\subsection{Experimental Settings}

% kafeng add from deepseek
\textbf{Training settings.}
Following SeeSR \cite{wu2024seesr} and OSEDiff \cite{li2024distillation}, we employ LSDIR \cite{li2023lsdir} and the first 10K images from FFHQ \cite{karras2019style} as the training data. We use the same degradation pipeline as RealESRGAN \cite{wang2021realesrgan} to synthesize LR-HR pairs. During training, the synthesized LR images are upscaled to match the HR resolution of $512 \times 512$ before feeding into our SR model. We train our model using the AdamW optimizer \cite{loshchilov2017decoupled} with a learning rate of $5\times10^{-5}$ on 4 A100 GPUs. The training process takes over 30k iterations, with a batch size of 16. We utilize the SD 2.1-base as the pretrained diffusion model for the $\times4$ SR task. The trainable LoRA modules \cite{hu2021lora} are applied to the weights of all convolution and MLP layers and are initialized using a Gaussian distribution. The rank of the LoRA utilized in our Expert-D4 training is set to 4. The rank in LoRA is set as 16 for the VAE encoder and 4 for the diffusion UNet in our Expert-D8 training, respectively.

\textbf{Testing Details.}
Following previous methods \cite{li2024distillation, wu2024seesr}, we evaluated our model on the synthetic DIV2K-Val \cite{agustsson2017ntire} dataset and two real-world datasets, RealSR \cite{cai2019realworld}, and DRealSR \cite{wei2020component}. The synthetic dataset consists of 3,000 image pairs, which are created by downgrading the $512\times512$ HQ images cropped from DIV2K-Val to $128\times128$ LQ images using the Real-ESRGAN \cite{wang2021realesrgan} degradation pipeline. The real-world datasets comprised $128\times128$ and $512\times512$ HQ image pairs. 

\textbf{Compared Methods.}
We categorize the test models into two groups: multi-step and single-step inference. We compare DDR-SR with leading multi-step DM-based methods StableSR \cite{wang2023stablesr}, DiffBIR \cite{lin2023diffbir}, SeeSR \cite{wu2024seesr}, PASD \cite{yang2023pixel}, and ResShift \cite{yue2024resshift}; one-step DM-based methods SinSR \cite{wang2023sinsr} and OSEDiff \cite{li2024distillation}.

\textbf{Evaluation Metrics.}
For the Real-ISR task, we utilize a set of reference-based and no-reference metrics to evaluate the competing methods. The reference-based metrics include PSNR, SSIM \cite{wang2004image}, LPIPS \cite{zhang2018unreasonable}, DISTS \cite{ding2020image}, and FID \cite{heusel2017gans}. Note that PSNR and SSIM, computed on the Y channel in the YCbCr space, are used to measure the fidelity of SR results; LPIPS and DISTS, computed in the RGB space, are used to assess the perceptual quality of SR results; FID  evaluates the distance of distributions between GT and restored images.
The no-reference metrics include CLIPIQA \cite{wang2023exploring}, MUSIQ \cite{ke2021musiq}, MANIQA \cite{yang2022maniqa} and NIQE \cite{zhang2015feature}.

\subsection{Comparison with State-of-the-Art Methods}

\textbf{Quantitative Comparisons.} 
 \Cref{tab:1} shows the quantitative comparison of our method with SD-based multi-step and single-step methods on three datasets for the Real-ISR task. Our DDR-SR achieves the best results in SSIM, LPIPS, and CLIPIQA metrics. Moreover, our method demonstrates highly competitive performance on both full-reference metrics (DIST and FID) and the no-reference metric MUSIQ, consistently ranking among the top two across all benchmarks. In terms of PSNR, DDR-SR outperforms all existing SD-based multi-step approaches, yet remains marginally behind ResShift and SinSR. DiffBIR, SeeSR, PASD, and DiffBIR exhibit better performance on the MANIQA and NIQE metrics, which may be attributed to the fact that multi-step models have more denoising iterations to produce rich details. However, our DDR-SR yields slightly inferior results on the MANIQA and NIQE, suggesting space for further improvement.
 
 \textbf{Qualitative Comparisons.} 
 \Cref{fig4} presents visual comparisons of different Real-ISR methods. It is evident that DRR-SR significantly outperforms other methods on structure preservation. Although the competing methods can improve image quality over the LR input, they struggle with preserving fine structures due to the high spatial compression ratio of the VAE, which discards high-frequency details during encoding. For instance, none of the compared methods were able to reconstruct the text in the first image or the pupil and eyelid textures in the second image. In contrast, our proposed DDR-SR method produces clearer and more legible text, as well as more realistic and detailed pupil and eyelid textures. 
 These results underscore DDR-SR’s superior detail fidelity.
 % These results underscore DDR-SR's superior capability in faithfully restoring high-fidelity image details. 
 More visualization comparisons can be found in the \textbf{supplementary materials}.

\begin{table}[!t]
\centering
\caption{Comparison of computational complexity across different SD-based Real-ISR methods.}
\label{tab:2}
\resizebox{1\linewidth}{!}{%
\begin{tabular}{l|ccccccc}
\hline
Methods & StableSR & DiffBIR & SeeSR & PASD & OSEDiff & Expert-D4 \\
\hline
Steps & 200 & 50 & 50 & 20 & 1 & 1 \\
Param. (B) & 1.56 & 1.68 & 2.51 & 2.31 & 1.77 & 1.63 \\
FLOPs (T) & 79.94 & 24.31 & 65.86 & 29.13 & 2.27 & 1.81 \\
\hline
\end{tabular}%
}
\end{table}

\textbf{Complexity Comparisons.}
 \Cref{tab:2} compares the SD-based Real-ISR methods in terms of parameter count (Para.)
and floating-point operations per second (FLOPs). Specifically,  Expert-D4 achieves the
lowest FLOPs at 1.81T, and it has 1.65B parameters, fewer than OSEDiff.

\subsection{Ablation Study}
To better understanding of the DDR strategy, and the roles of the LoRA modules. We conduct ablation studies to investigate the impact of the LoRA rank configuration in both the VAE and the diffusion UNet of our expert networks. \Cref{tab:3,tab:4} show the impact of varying the LoRA rank on Expert-D4 and Expert-D8, respectively. We adopt LoRA ranks $\{4,4\}$ for Expert-D4 and $\{16,4\}$ for Expert-D8.

\begin{table}[!t]
\centering
\caption{Ablation study on LoRA rank configurations in Expert-D4 on the RealSR dataset.}
\label{tab:3}
\resizebox{1\linewidth}{!}{%
\begin{tabular}{cc|cccc}
\hline
Rank of VAE-D4 & Rank of UNet & PSNR$\uparrow$ & SSIM$\uparrow$ & CLIPIQA$\uparrow$ & MUSIQ$\uparrow$  \\
\hline
4 &4    & 26.04   & 0.7560    & 0.6996    & 70.12  \\
8 &4    & 26.04  & 0.7580  & 0.6944  & 70.19 \\
16 &4    & 25.91 & 0.7567 & 0.6873 & 69.81 \\
8 &8    & 26.19 & 0.7584 & 0.6748 & 70.18 \\
8 &16    & 26.16 & 0.7604 & 0.6880 & 70.03 \\
\hline
\end{tabular}%
}
\end{table}

\begin{table}[!t]
\centering
\caption{Ablation study on LoRA rank configurations in Expert-D8 on the RealSR dataset.}
\label{tab:4}
\resizebox{1\linewidth}{!}{%
\begin{tabular}{cc|cccc}
\hline
Rank of VAE-D8 & Rank of UNet & PSNR$\uparrow$ & SSIM$\uparrow$ & CLIPIQA$\uparrow$ & MUSIQ$\uparrow$  \\
\hline
4 &4    & 25.86   & 0.7587    & 0.6414    & 68.16  \\
16 &4    & 25.78  & 0.7550  & 0.6454  & 68.51 \\
8 &8    & 26.26 & 0.7618 & 0.6237 & 67.54 \\
\hline
\end{tabular}%
}
\end{table}

\section{Conclusion}

% kafeng add from deepseek
% We propose DDR-SR, a difficulty-aware dynamic routing framework for real-world image super-resolution. By estimating restoration difficulty and adaptively routing inputs to specialized experts, DDR-SR achieves an effective balance between reconstruction quality and inference efficiency. Experiments show that our method attains state-of-the-art perceptual performance with only one denoising step, while offering user-controllable trade-offs between fidelity and speed. The framework demonstrates the promise of adaptive routing for efficient high-quality image restoration.

% This paper introduces DDR-SR, a difficulty-aware dynamic routing framework for real-world image super-resolution. DDR-SR assesses input difficulty via high-frequency energy and adaptively routes samples to two specialized experts: Expert-D4 for challenging cases and Expert-D8 for simpler inputs, both efficiently fine-tuned from a pre-trained diffusion model via LoRA. Experiments demonstrate that DDR-SR achieves state-of-the-art perceptual quality with only a single denoising step, while providing tunable control over the quality-efficiency trade-off. These results validate the effectiveness of difficulty-aware routing in balancing super-resolution performance and inference efficiency. Future work may extend the routing mechanism to multiple experts and apply it to related image restoration tasks.

We present DDR-SR, a difficulty-aware dynamic routing framework for real-world image super-resolution. It adaptively routes inputs between two specialized experts based on a high-frequency energy-based difficulty estimator: Expert-D4 for high-fidelity restoration of hard samples, and Expert-D8 for efficient processing of easy ones. Through mixed training and LoRA-based tuning, DDR-SR achieves complementary expertise and efficient inference.
Experiments show DDR-SR outperforms state-of-the-art methods across synthetic and real-world benchmarks, offering a tunable trade-off between quality and speed. By moving beyond one-size-fits-all processing, DDR-SR provides a more adaptive and practical solution for Real-ISR.

\section*{Impact Statement}

This paper presents work whose goal is to advance the field of Machine
Learning. There are many potential societal consequences of our work, none
of which we feel must be specifically highlighted here.

% In the unusual situation where you want a paper to appear in the references without citing it in the main text, use \nocite

\bibliography{paper}
\bibliographystyle{icml2026}

%%%%%%%%%%%%%%%%%%%%%%%%%%%%%%%%%%%%%%%%%%%%%%%%%%%%%%%%%%%%%%%%%%%%%%%%%%%%%%%
%%%%%%%%%%%%%%%%%%%%%%%%%%%%%%%%%%%%%%%%%%%%%%%%%%%%%%%%%%%%%%%%%%%%%%%%%%%%%%%
% APPENDIX
%%%%%%%%%%%%%%%%%%%%%%%%%%%%%%%%%%%%%%%%%%%%%%%%%%%%%%%%%%%%%%%%%%%%%%%%%%%%%%%
%%%%%%%%%%%%%%%%%%%%%%%%%%%%%%%%%%%%%%%%%%%%%%%%%%%%%%%%%%%%%%%%%%%%%%%%%%%%%%%
\newpage
\appendix
\onecolumn
\section{Visualization Comparisons}
\begin{figure*}[h]
  \begin{center}
    \centerline{\includegraphics[width=\linewidth]{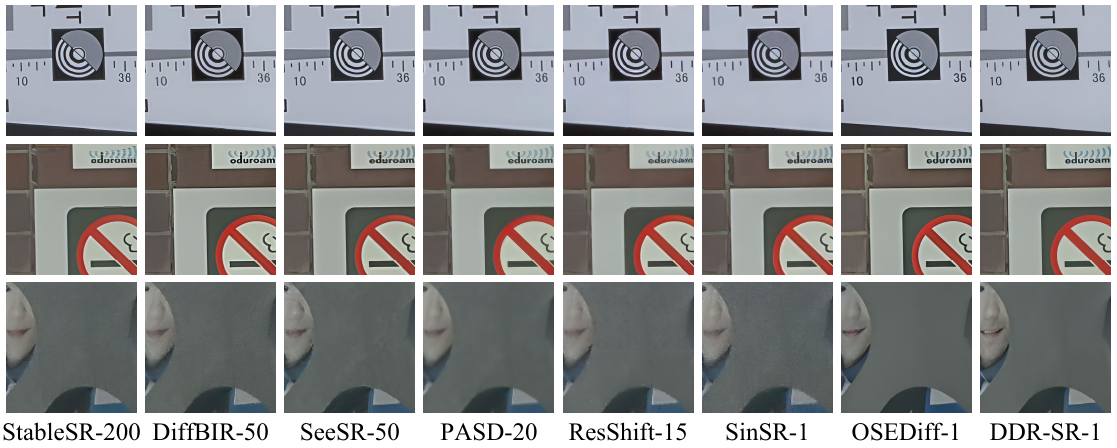}}
    \caption{
    Visual comparisons of different Real-ISR methods. Please zoom in for a better view.
    }
    \label{fig:5}
  \end{center}
  \vskip -0.2in
\end{figure*}

\Cref{fig:5} provides a visual comparison of various diffusion-based methods applied to real-world datasets, covering both multi-step and one-step approaches. As shown, none of the competing methods could effectively reconstruct the intricate ruler texture in the first image, the letters and graphics in the second image, or the facial details in the third image. Specifically, these methods either produced blurred or incomplete reconstructions, failing to capture the fine structural details.

In contrast, our DDR-SR method excels in these challenging scenarios. It successfully reconstructs the ruler's texture with sharp edges and minimal artifacts, renders the letters and graphics clearly and accurately, and captures detailed facial features, including subtle details like beard stubble. These results highlight DDR-SR's superior capability in faithfully restoring high-fidelity image details, demonstrating its effectiveness in handling diverse and complex real-world degradations.
% You can have as much text here as you want. The main body must be at most $8$
% pages long. For the final version, one more page can be added. If you want, you
% can use an appendix like this one.

% The $\mathtt{\backslash onecolumn}$ command above can be kept in place if you
% prefer a one-column appendix, or can be removed if you prefer a two-column
% appendix.  Apart from this possible change, the style (font size, spacing,
% margins, page numbering, etc.) should be kept the same as the main body.
%%%%%%%%%%%%%%%%%%%%%%%%%%%%%%%%%%%%%%%%%%%%%%%%%%%%%%%%%%%%%%%%%%%%%%%%%%%%%%%
%%%%%%%%%%%%%%%%%%%%%%%%%%%%%%%%%%%%%%%%%%%%%%%%%%%%%%%%%%%%%%%%%%%%%%%%%%%%%%%

\end{document}